\documentclass[]{ailab}
\newcommand{\ours}{\texttt{Free()LM}}

\usepackage{wrapfig}
\usepackage{tabularx}
\usepackage{textcomp}
\usepackage{stfloats}
\usepackage{url}
\usepackage{verbatim}
\usepackage{titlesec}
\usepackage{tocloft}
\usepackage{adjustbox}
\usepackage{multirow}
\usepackage{pifont}
\usepackage{tikz}
\usepackage{comment}
\usepackage{amsmath,amssymb} 
\usepackage{colortbl}  
\usepackage{color}
\usepackage{booktabs} 
\usepackage{natbib} 
\setcitestyle{square,comma,numbers,sort&compress}
\usepackage{graphicx}    
\usepackage{subcaption} 
\usepackage{multirow} 
\usepackage{tcolorbox}
\usepackage{booktabs} 
\usepackage{subcaption} 
\usepackage[dvipsnames]{xcolor}
\usepackage{graphicx}
\usepackage{amsmath, amssymb}
\RequirePackage{xspace}
\makeatletter
\DeclareRobustCommand\onedot{\futurelet\@let@token\@onedot}
\def\@onedot{\ifx\@let@token.\else.\null\fi\xspace}

\definecolor{cotred}{RGB}{200, 30, 30}
\definecolor{ForestGreen}{RGB}{34, 139, 34}

\newcommand{\cotblock}[2]{%
  \par\noindent
  \colorbox{#1}{\parbox{0.96\linewidth}{#2}}%
  \par
}

\usepackage[normalem]{ulem}
\newcommand{\cotdel}[1]{\textcolor{red}{\sout{#1}}}

\newcommand{\cotgen}[1]{\textcolor{ForestGreen}{#1}}

\makeatother

\definecolor{adptorange}{RGB}{248, 205, 172}
\definecolor{cmpblue}{RGB}{189, 215, 238}
\definecolor{cmpblue}{RGB}{189, 215, 238}

\definecolor{our_red}{RGB}{232,157,160}
\definecolor{our_blue}{RGB}{136,206,230}
\definecolor{our_orange}{RGB}{246,200,168}
\definecolor{our_green}{RGB}{178,211,164}

\definecolor{attn_code0}{RGB}{247,215,200}
\definecolor{attn_code1}{RGB}{238,169,139}
\definecolor{mlp_code0}{RGB}{204,201,221}
\definecolor{mlp_code1}{RGB}{102,95,153}

\definecolor{token_blue}{RGB}{84, 120, 140}

\usepackage{amsmath,amsfonts,bm}

\def\eqref#1{equation~\ref{#1}}

\def\1{\bm{1}}

\DeclareMathAlphabet{\mathsfit}{\encodingdefault}{\sfdefault}{m}{sl}
\SetMathAlphabet{\mathsfit}{bold}{\encodingdefault}{\sfdefault}{bx}{n}

\usepackage{amsmath,amsfonts,bm}

\def\eqref#1{equation~\ref{#1}}

\def\1{\bm{1}}

\DeclareMathAlphabet{\mathsfit}{\encodingdefault}{\sfdefault}{m}{sl}
\SetMathAlphabet{\mathsfit}{bold}{\encodingdefault}{\sfdefault}{bx}{n}

\usepackage{multirow}
\usepackage{diagbox}
\usepackage{makecell}
\usepackage{tabularx}
\usepackage{graphicx}

\usepackage{array}
\usepackage{rotating}

\definecolor{aliceblue}{rgb}{0.94, 0.97, 1.0}
\definecolor{citecolor}{HTML}{0071BC}
\definecolor{linkcolor}{HTML}{ED1C24}
\definecolor{darkgreen}{HTML}{539165}

\makeatletter
\newcommand{\thickhline}{%
 \noalign {\ifnum 0=`}\fi \hrule height 1pt
 \futurelet \reserved@a \@xhline
}
\makeatother

\usepackage{pifont}

\usepackage{pifont}       
\usepackage{bbding}       
\usepackage{fontawesome}
\usepackage{xspace}

\usepackage{float}
\usepackage{enumitem}

\newlength\savewidth

\newcolumntype{x}[1]{>{\centering\arraybackslash}p{#1pt}}
\newcolumntype{y}[1]{>{\raggedright\arraybackslash}p{#1pt}}
\newcolumntype{z}[1]{>{\raggedleft\arraybackslash}p{#1pt}}

\renewcommand{\paragraph}[1]{\vspace{1mm}\noindent\textbf{#1}}
\usepackage{colortbl}
\usepackage{xcolor}
\usepackage{wrapfig}

\renewcommand{\paragraph}[1]{\vspace{1.25mm}\noindent\textbf{#1}}

\usepackage{algorithm}
\usepackage{listings}

\definecolor{codeblue}{rgb}{0.25, 0.5, 0.5}
\definecolor{codekw}{rgb}{0.35, 0.35, 0.75}
\lstdefinestyle{Pytorch}{
    language = Python,
    backgroundcolor = \color{white},
    basicstyle = \fontsize{9pt}{8pt}\selectfont\ttfamily\bfseries,
    columns = fullflexible,
    aboveskip=1pt,
    belowskip=1pt,
    breaklines = true,
    captionpos = b,
    commentstyle = \color{codeblue},
    keywordstyle = \color{codekw},
}

\definecolor{colSubject}{HTML}{D32F2F}   
\definecolor{colAction}{HTML}{F57C00}    
\definecolor{colDetail}{HTML}{388E3C}    
\definecolor{colSpatial}{HTML}{1976D2}   
\definecolor{colMood}{HTML}{7B1FA2}      
\definecolor{colKnow}{HTML}{AFB42B}      

\usepackage{xcolor}

\usepackage{tikz}
\usetikzlibrary{tikzmark, calc, shadows.blur, shapes.geometric, fit, positioning}
\usepackage{xparse}
\pgfdeclarelayer{background}
\pgfdeclarelayer{floatbox}
\pgfdeclarelayer{foreground}
\pgfsetlayers{background,floatbox,main,foreground}

\newcounter{hcellcount}

\NewDocumentCommand{\hctext}{m}{\csname hctext@#1\endcsname}
\NewDocumentCommand{\sethctext}{mm}{\expandafter\gdef\csname hctext@#1\endcsname{#2}}

\definecolor{scoreRed}{RGB}{200, 0, 0}
\definecolor{grayText}{RGB}{120, 120, 120}

\definecolor{green}{HTML}{009000}
\definecolor{red}{HTML}{ea4335}

\definecolor{cvblue}{rgb}{0.15, 0.45, 0.68}

\title{\centering Free(): Learning to Forget in Malloc-Only Reasoning Models}

\author[1, 2, *]{Yilun Zheng}
\author[1,*]{Dongyang Ma}
\author[1, *]{Tian Liang}
\author[1]{Jiahao Xu}
\author[1]{Xinting Huang}
\author[2]{Lihui Chen}
\author[1]{Haitao Mi}
\author[1, *, \dagger]{Yan Wang}

\affiliation[1]{Tencent AI Lab\\}
\affiliation[2]{Nanyang Technological University}
\contribution[*]{Equal contribution}
\contribution[\dagger]{Project Lead}

\email{yanwang.branden@gmail.com }

\abstract{
Reasoning models enhance problem-solving by scaling test-time compute, yet they face a critical paradox: excessive thinking tokens often degrade performance rather than improve it. We attribute this to a fundamental architectural flaw: standard LLMs operate as \textbf{``malloc-only'' engines}, continuously accumulating valid and redundant steps alike without a mechanism to prune obsolete information. To break this cycle, we propose \textbf{\ours{}}, a model that introduces an intrinsic self-forgetting capability via the \textbf{Free-Module}, a plug-and-play LoRA adapter. By iteratively switching between reasoning and cleaning modes, \ours{} dynamically identifies and prunes useless context chunks, maintaining a compact and noise-free state.

Extensive experiments show that Free()LM provides consistent improvements across all model scales (8B to 685B). It achieves a 3.3\% average improvement over top-tier reasoning baselines, even establishing a new \textbf{SOTA} on IMOanswerBench using DeepSeek V3.2-Speciale.
Most notably, in long-horizon tasks where the standard Qwen3-235B-A22B model suffers a total collapse (0\% accuracy), \ours{} restores performance to \textbf{$\sim$50\%}. Our findings suggest that sustainable intelligence requires the freedom to forget as much as the power to think.
}

\date{\today} 
\headercontent{
    \begin{tabular}{ccc}
        \href{https://github.com/TemporaryLoRA/FreeLM}{\faGithub\ Code} &
        \hspace{2em}
        
        \href{https://hf.co/collections/ldsjmdy/freelm}{%
            \raisebox{-0.15\height}{\includegraphics[height=1.1em]{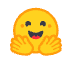}}\ Models%
        } & 
        \hspace{2em}
        
        \href{https://huggingface.co/datasets/ldsjmdy/FreeLM}{%
            \faDatabase\ Data
        }
    \end{tabular}
}

\begin{document}
\thispagestyle{firstheader}
\maketitle

\section{Introduction}
\label{sec:intro}

\begin{wrapfigure}{r}{0.5\textwidth} 
  \centering
  \vspace{-10pt} 
  \includegraphics[width=\linewidth]{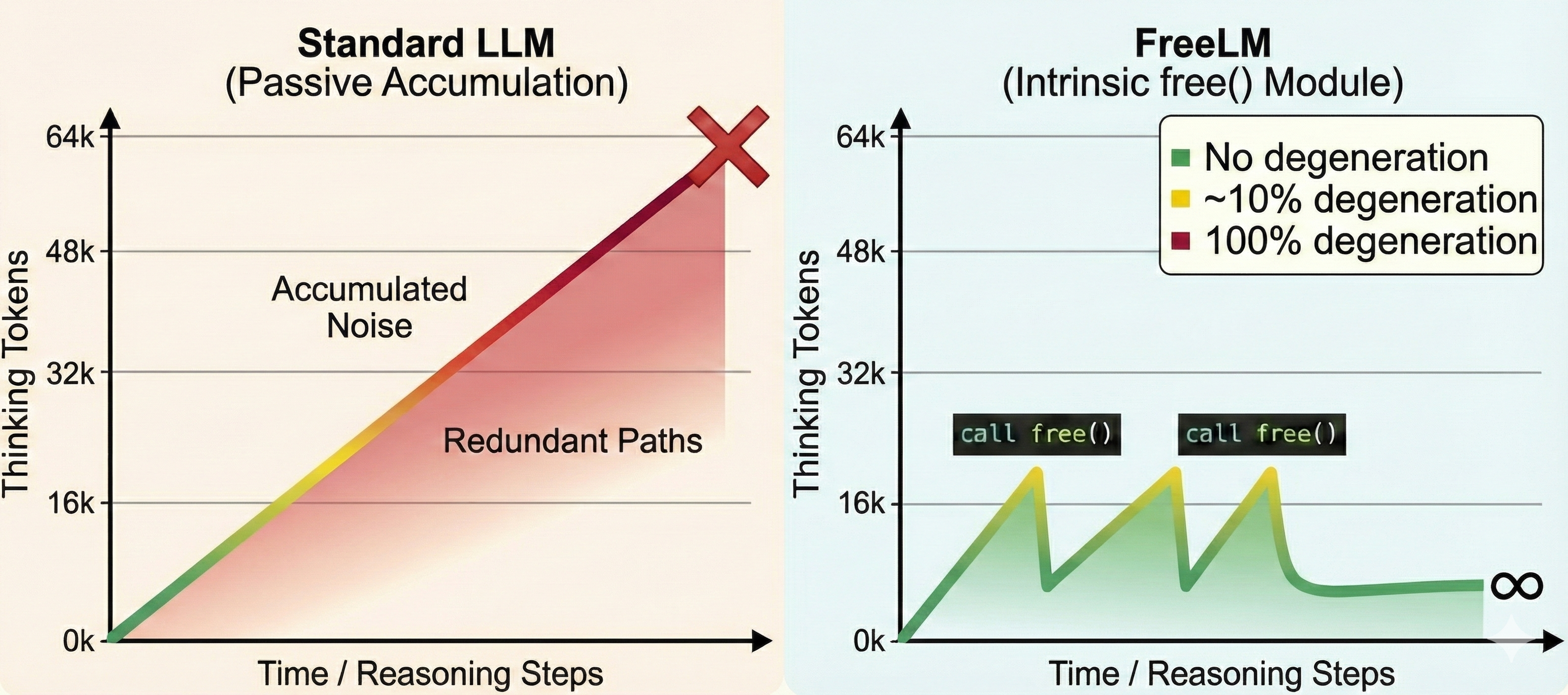} 
  \vspace{5pt}
  \caption{
    \textbf{Empirical observation of Qwen3-8B reasoning on AIME benchmarks.}
    \textbf{Left:} Standard LLMs passively accumulate tokens, causing the reasoning process to eventually ``crash'' (degenerate).
    \textbf{Right:} \ours{} integrates an intrinsic \texttt{free()} mechanism. By periodically identifying and pruning redundant reasoning steps, it actively maintains a compact, noise-free state, enabling sustainable long-chain reasoning.
  }
  \label{fig:concept}
  \vspace{-20pt} 
\end{wrapfigure}

Reasoning models have unlocked powerful problem-solving skills by scaling test-time compute—using more ``thinking tokens'' to reason before answering. However, this approach faces a critical bottleneck: performance does not simply improve with longer thoughts. Instead, excessive reasoning often causes accuracy to drop~\citep{contextrot} or even leads to severe degeneration (e.g., falling into repetitive loops). Consequently, the utility of test-time scaling is bounded: empirically, if a solution is not reached once thinking tokens occupy the majority of the context window (approx. 70--90\%), further reasoning typically yields no benefit.

We validate this bottleneck in Figure~\ref{fig:concept} (Left) using Qwen3-8B, a reasoning model with a 32k context window. Across 480 reasoning trajectories sampled from AIME 24\&25~\citep{aime24, aime25}, degeneration rises sharply once thinking tokens exceed 16k. Specifically, among the 31 instances reaching the context limit, 26 (84\%) were already trapped in repetitive loops, and the degeneration ratio hits 100\% by 48k, causing a total reasoning collapse. This phenomenon reveals an architectural flaw: standard reasoning models operate as ``malloc-only'' engines. They passively accumulate every step without a \texttt{free()} mechanism to discard useless information. Consequently, instead of boosting performance, we face a paradox: with greater thinking comes not greater power, but greater noise.

To break this ``malloc-only'' cycle, we introduce \textbf{\ours{}}. Our design stems from a simple observation: complex reasoning produces temporary waste, such as intermediate steps or trial paths that become obsolete once resolved. \ours{} integrates an intrinsic \texttt{free()} mechanism to periodically prune these redundancies. Specifically, such a self-forget capability is implemented by plugging in an additional LoRA adapter, which we term the \textbf{Free-Module}. By dynamically merging and unmerging this module, \ours{} switches between two modes: (1) \textbf{Reasoning mode (unmerged)}, where the model remains equivalent to the original backbone, focusing purely on generating next tokens to solve the problem; and (2) \textbf{Cleaning mode (merged)}, where the module activates to scan the context, identify useless segments, and output explicit commands (specifying a \texttt{prefix} and \texttt{suffix}) to prune them. By dynamically switching between these two modes, the model continuously maintains a compact, noise-free state for sustainable reasoning.

We achieve a \textbf{3.3\%} average gain across six reasoning benchmarks (e.g., IMOanswerBench~\citep{IMOAnswer}, HLE~\citep{HLE}) for reasoning models ranging from \textbf{8B to 685B}. Specifically, \ours{} pushes DeepSeek V3.2-Speciale to a \textbf{new SOTA} on IMOanswerBench with a \textbf{2.3\%} accuracy boost. More importantly, \ours{} solves the collapse of long-horizon reasoning: while the standard Qwen3-235B drops to \textbf{0\% accuracy} on complex tasks requiring $>$80k thinking tokens, \ours{} recovers performance to \textbf{$\sim$50\%}. These results prove our core point: sustainable intelligence requires the freedom to forget as much as the power to think.

In summary, our contributions are as follows: 

\begin{itemize}
    \item We propose \ours{} to break the ``malloc-only'' limit of current models. Using a plug-and-play LoRA adapter (the \textbf{Free-Module}), \ours{} enables models to prune redundant context and maintain a compact, noise-free state.

    \item We design an efficient pruning method. By generating only prefixs and suffixs, the model can remove large redundant chunks with minimal cost.

    \item We validate \ours{} on six benchmarks using top-tier models from \textbf{8B up to 685B}. It achieves a 3.3\% average boost and sets a \textbf{new SOTA} on IMOanswerBench with DeepSeek V3.2-Speciale.
\end{itemize}

\section{Method}
\label{sec:method}

\begin{figure*}[t]
  \centering
  \includegraphics[width=\linewidth]{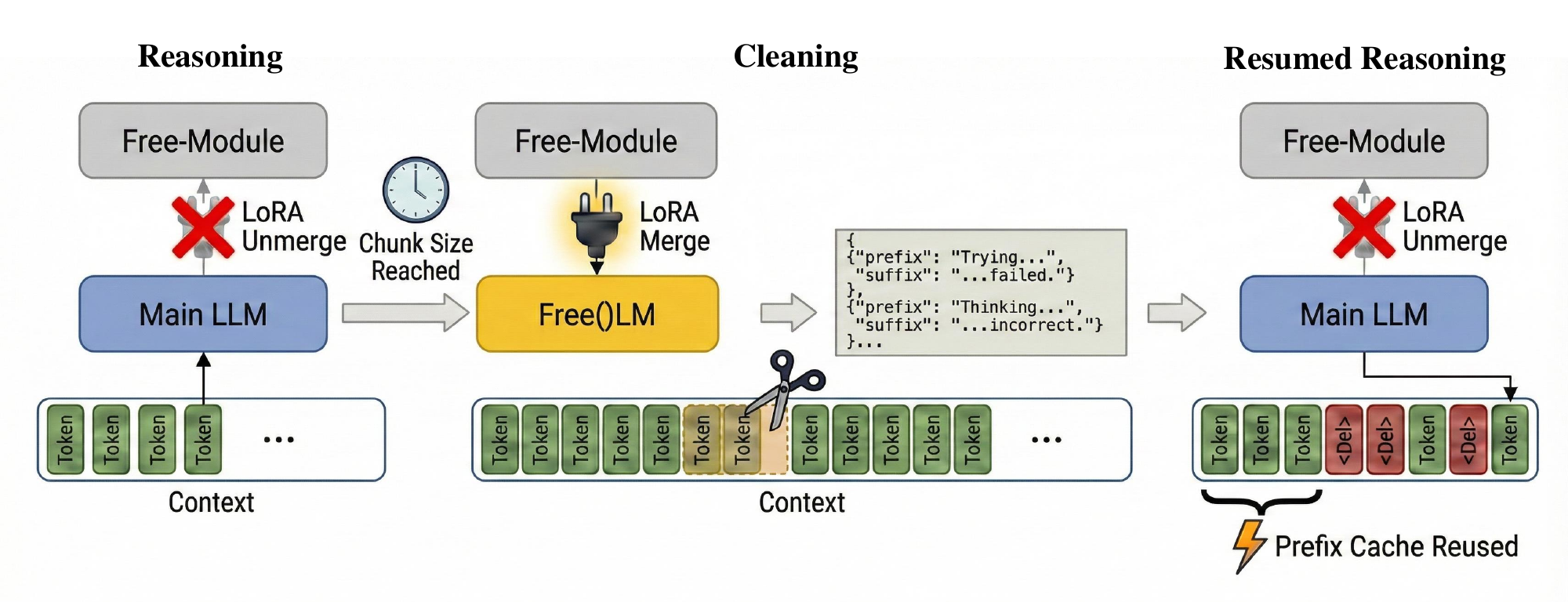}
  \caption{\textbf{The \textbf{\ours{}} Inference Framework.} The model operates on a cyclic Reasoning-Cleaning mechanism. \textbf{Reasoning:} The Main model generates tokens normally with the Free-Module unmerged. \textbf{Cleaning:} Upon reaching a chunk limit, the Free-Module is merged to identify and prune redundant chunks. \textbf{Resumed Reasoning:} The module is unmerged, and reasoning resumes on the cleaned context.}
  \label{fig:main_architecture}
\end{figure*}

To address the aforementioned ``malloc-only'' limitation of standard reasoning models, we propose \ours{}, a novel architecture that augments the LLM backbone with a lightweight, plug-and-play LoRA adapter,  Free-Module. As illustrated in Figure~\ref{fig:main_architecture}, by dynamically merging and unmerging this module, \ours{} switches between two distinct modes:

\begin{itemize}
    \item \textbf{Reasoning (Unmerged):} In this mode, the model remains equivalent to the original backbone, focusing exclusively on generating reasoning tokens to solve the problem.
    \item \textbf{Cleaning (Merged):} Upon activation, the Free-Module is merged into the backbone. The model shifts its focus to memory management, scanning the context to identify redundancies and outputting pruning commands.
\end{itemize}

In the following subsections, we detail the \ours{} architecture in \S\ref{sec:inference} and the training methodology in \S\ref{sec:training}.


\subsection{Architecture \& Inference}
\label{sec:inference}

Structurally, the Free-Module is a LoRA adapter~\citep{hu2021loralowrankadaptationlarge} attached to the standard LLM backbone. Once merged (Step 2 in Figure~\ref{fig:main_architecture}), \ours{} switches to \textbf{Cleaning Mode}. In this state, the model halts reasoning and scans the existing context to identify redundancies. It outputs a structured \textbf{pruning command} in JSON format:
\[
\texttt{[\{"prefix": "...", "suffix": "..."\},...]}
\]
Here, \texttt{prefix} and \texttt{suffix} serve as unique string anchors defining the span $[\text{start}, \text{end}]$ to be pruned. This anchor-based design enables the module to efficiently prune very long chunks of redundancy by generating only a few command tokens.

\paragraph{Pruning \& Resumed Inference:}
Upon receiving the command, an external Python executor parses the JSON and prunes the targeted spans via string matching:
\[
\texttt{context = re.sub(prefix + r'.*?' + suffix, '<Del>', context)}
\]
Following pruning, the \ours{} unmerges the Free-Module to revert to Reasoning Mode. To resume generation on the cleaned context, we identify two strategies:
\begin{enumerate}
    \item \textbf{Re-prefilling.} We reuse the KV cache for the unchanged prefix and strictly re-prefill the altered suffix.
    \item \textbf{KV Cache Pruning.} We directly excise the deleted memory blocks. To address the positional shift of subsequent tokens, we rotate their cached Key-Value states to re-align with the correct position indices.~\cite{mablock}
\end{enumerate}

Our pilot experiments on smaller models indicate that both strategies yield nearly identical performance. However, since \textbf{Strategy 2} is not natively supported by standard serving frameworks like vLLM \citep{vllm}, we adopt \textbf{Strategy 1} for our main experiments to ensure compatibility and efficiency.

\paragraph{Triggering:}
We introduce a hyperparameter, the \textbf{Pruning Interval} ($L_{\text{clean}}$). The model simply merges the Free-Module to trigger a cleaning cycle for every $L_{\text{clean}}$ reasoning tokens generated.

\begin{figure}[t]
  \centering
  \includegraphics[width=1\linewidth]{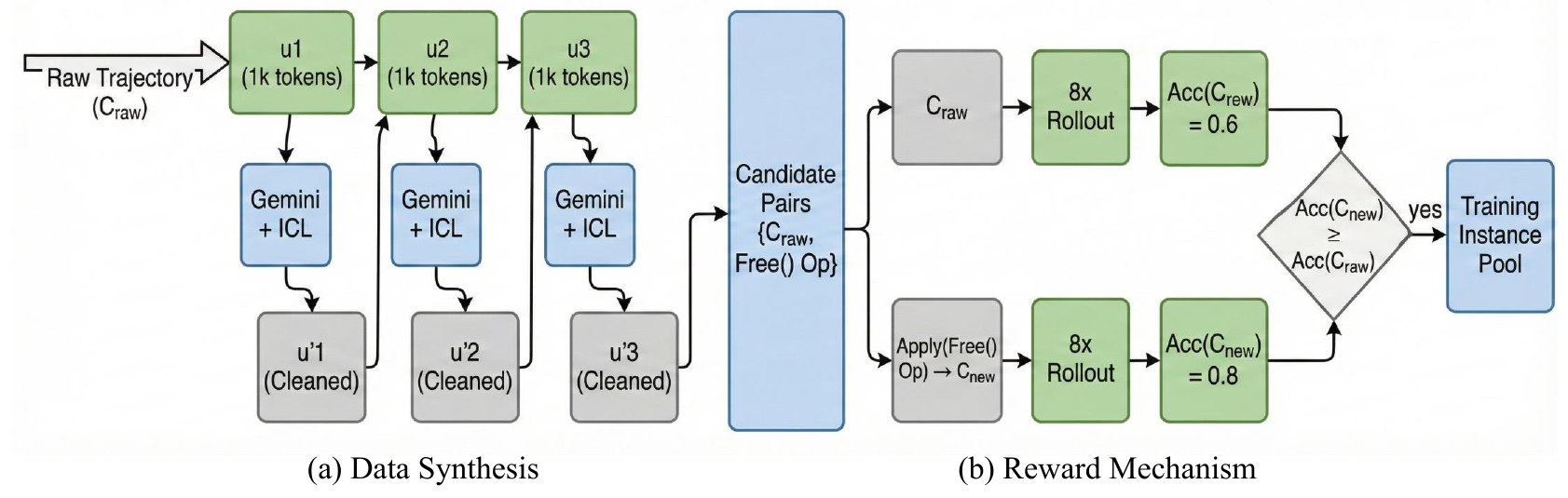}
\caption{
  \textbf{The Data Construction Pipeline.}
  \textbf{(a) Data Synthesis:} We segment raw trajectories into 1k-token chunks and employ Gemini-2.5-Pro to \textbf{sequentially} generate candidate training instances.
  \textbf{(b) Reward Mechanism:} By executing $K=8$ parallel rollouts, we retain an instance only if the pruned context $C_{\text{new}}$ maintains or improves accuracy compared to the original ($\text{Acc}(C_{\text{new}}) \ge \text{Acc}(C_{\text{raw}})$).
}
  \label{fig:data}
\end{figure}

\subsection{Training: Learning to Forget}
\label{sec:training}
Given the concept of active context management, a natural starting point is to explore In-Context Learning (ICL) solutions: asking the model itself to conduct self-correction or employing a strong external LLM to identify redundancy. Unfortunately, our preliminary experiments (detailed in Table~\ref{tab:qwen3-performance}) reveal the severe limitations of this approach. Even with extensive prompt optimization and utilizing powerful models like Gemini-2.5-Pro \citep{gemini} as experts, the performance gains on Qwen3-8B were marginal ($\sim1\%$). This finding suggests that effectively executing the \texttt{free()} operation—distinguishing necessary historical context from obsolete noise—is a complex capability that must be acquired through explicit training.

To address the lack of ground-truth labels, we propose a data pipeline centered on a sophisticated reward mechanism. Specifically, we first synthesize a large pool of candidate \texttt{free()} operations via ICL solutions, and subsequently filter for high-quality instances using the reward mechanism. The overall pipeline is illustrated in Figure~\ref{fig:data}.

\paragraph{Data Synthesis}
We randomly select 1,000 trajectories from DeepMath-103k~\cite{he2025deepmath103klargescalechallengingdecontaminated} and synthesize supervision signals using Gemini-2.5-Pro.
Crucially, to mimic the run-time behavior, we adopt a sequential pruning strategy. We segment each raw trajectory $T$ into sequential 1k-token chunks $\{u_1, u_2, \dots, u_n\}$. The cleaning process is performed iteratively: when the oracle pruns the current chunk $u_k$, it is conditioned on the already cleaned history prefix $H'_{k-1}$ (composed of previous cleaned chunks) rather than the original raw context. This dependency chain ensures that the training data reflects the fragmented memory states the model will actually encounter during inference. This pipeline yielded $\sim$8,000 candidate instances.

\paragraph{Reward Mechanism}
To filter the synthesized candidates, we employ a rigorous rejection sampling strategy. We treat the dataset as pairs of $\{C_{\text{raw}}, \mathcal{O}\}$, where $C_{\text{raw}}$ is the original context and $\mathcal{O}$ denotes the candidate pruning command.

For each pair, we execute 8 independent reasoning rollouts on both the original context $C_{\text{raw}}$ and the cleaned context $C_{\text{new}} = \text{Apply}(C_{\text{raw}}, \mathcal{O})$. A candidate operation is retained if and only if the pruning preserves or improves the accuracy:
\[
\text{Acc}(C_{\text{new}}) \ge \text{Acc}(C_{\text{raw}})
\]
This strict criterion ensures that the module learns to prune noise without reducing the probability of reaching the correct solution. Consequently, this verification process distilled the dataset down to 6,648 high-quality training instances.
\section{Experiments}

We evaluate \ours{} across a diverse set of benchmarks to verify its effectiveness, generalization capability, and efficiency. Our experiments cover models ranging from 8B to 685B parameters.

\subsection{Settings}


\paragraph{Backbone Models.} We evaluate \ours{} across three model scales: Qwen3-8B (thinking mode)\footnote{\url{https://huggingface.co/Qwen/Qwen3-8B}}, Qwen3-30B-A3B-Thinking-2507\footnote{\url{https://huggingface.co/Qwen/Qwen3-30B-A3B-Thinking-2507}}, and Qwen3-235B-A22B-Thinking-2507\footnote{\url{https://huggingface.co/Qwen/Qwen3-235B-A22B-Thinking-2507}} (hereafter denoted as \textbf{Qwen3-8B}, \textbf{Qwen3-30B-A3B}, and \textbf{Qwen3-235B-A22B} for brevity). We set the context window to 32k for Qwen3-8B and Qwen3-30B-A3B, as most reasoning paths fit within this limit. For the largest model, Qwen3-235B-A22B, we allocate a 128k window for the \textbf{Vanilla} baseline to accommodate its extensive generation length, while restricting \ours{} to 64k, leveraging its ability to actively free up memory during inference.

\paragraph{Compared Methods.}
We compare our \ours{} with three types of baselines:
\begin{itemize}
    \item \textbf{Vanilla}: The standard backbone model, \textbf{Qwen3-8B}, \textbf{Qwen3-30B-A3B}, and \textbf{Qwen3-235B-A22B} \citep{qwen3} without any context management or compression techniques.

    \item \textbf{Heuristic Compression}: Since KV cache compression naturally results in a shorter context, we adapt \textbf{H2O}~\citep{h2o} and ThinkCleary (\textbf{TC})~\citep{thinkclearly} as heuristic baselines. We apply these methods dynamically during reasoning to prune tokens based on their accumulated attention scores. Note that due to their incompatibility with the vLLM \citep{vllm} serving framework, we evaluate them exclusively on Qwen3-8B.
    
    
    \item \textbf{ICL Methods}: the ICL solutions discussed in Section~\ref{sec:training}.  We instruct both the backbone models (denoted as \textbf{No Train}) and the SOTA \textbf{Gemini}-2.5-Pro \citep{gemini} to execute \texttt{free()} operations via a sophisticatedly optimized prompt detailed in Appendix~\ref{sec:prompts}. 
    

\end{itemize}

\paragraph{Benchmarks.}
To comprehensively evaluate \ours{}, we curate a diverse benchmark suite designed to stress-test two critical hypotheses: (1) \textit{whether the learned active forgetting capability effectively enhances long-horizon reasoning as intended}; and (2) \textit{whether the \texttt{free()} mechanism is safe, preserving performance on general tasks where extensive pruning is unnecessary}.

To assess the first hypothesis (Long-Horizon Reasoning), we employ a suite of challenging benchmarks: AIME 24\&25~\citep{aime24, aime25},\footnote{AIME24~\citep{aime24} and AIME25~\citep{aime25} are combined into a single dataset, hereafter referred to as AIME2425.} BrUMO25~\citep{brumo25}, HMMT~\citep{hmmt}, BeyondAIME~\citep{beyondaime}, Human's Last Exam (HLE, text-only subset)~\citep{HLE}, and IMOAnswerBench~\citep{IMOAnswer}. We selected these datasets for two primary reasons. First, they demand complex, multi-step logic that typically yields trajectories exceeding 20k tokens—a regime where standard models frequently suffer from catastrophic degeneration. Second, the recency of datasets like BeyondAIME (June 2025) and IMOAnswerBench (November 2025) minimizes data contamination risks, ensuring a robust evaluation of intrinsic reasoning capabilities.

To assess the second hypothesis (General Reasoning \& Safety), we evaluate \ours{} on standard benchmarks requiring shorter reasoning chains: BBH~\citep{BBH}, MMLU-Pro~\citep{MMLU_pro}, MMLU-STEM~\citep{MMLU_stem}, and GPQA-Diamond~\citep{gpqa}. These tasks serve as a control group to confirm that \ours{} maintains the baseline performance of standard models on general-purpose queries.

\paragraph{Implementation Details.}
For mathematical reasoning tasks, we standardize outputs using the prompt: ``\texttt{Please reason step by step, and put your final answer within $\backslash$boxed\{\}.}'' Correctness is evaluated by extracting the answer from ``\texttt{$\backslash$boxed\{\}}'' and matching it against the ground truth, following the evaluation pipeline \citep{he2025deepmath103klargescalechallengingdecontaminated}. We report three key metrics: pass@1, the number of response tokens (\#Token), and the response reduction ratio ($\Delta$) relative to the vanilla model. For all experiments, we employ sampling-based decoding with temperature=0.7, top\_k=20, and top\_p=0.95. To balance computational cost with statistical reliability, we configure the number of rollouts based on dataset size: 8 rollouts for AIME2425, BrUMO25, HMMT, and BeyondAIME (which have limited questions), and 1 rollout for the larger HLE and IMOAnswerBench datasets. For \ours{}, the pruning interval $L_\text{clean}$ is set to 5000 with a maximum of 50 iterations.

\subsection{Main Results}
\label{sec:exp_main}
We begin by addressing the core question: Can \ours{} enhance reasoning performance through active pruning? As shown in Table~\ref{tab:qwen3-performance}, \ours{} consistently improves accuracy across all benchmarks while maintaining a significantly more compact context. 

\begin{table*}[tbp]
\centering
\caption{Performance of Qwen3 models. We report pass@1 (p@1) performance computed over 8 rollouts, along with the average number of response tokens (\#Token). For the Average columns, brackets represent the absolute change for p@1 and the relative change for \#Token (where blue indicates improvement and red indicates regression).}
\label{tab:qwen3-performance}
\resizebox{\textwidth}{!}{
\begin{tabular}{llcccccccccccccc}
\toprule
\multirow{2}{*}{\textbf{Model}} & \multirow{2}{*}{\textbf{Setting}} &
\multicolumn{2}{c}{\textbf{AIME2425}} &
\multicolumn{2}{c}{\textbf{BrUMO25}} &
\multicolumn{2}{c}{\textbf{HMMT}} &
\multicolumn{2}{c}{\textbf{BeyondAIME}} &
\multicolumn{2}{c}{\textbf{HLE}} &
\multicolumn{2}{c}{\textbf{IMOAnswer}} &
\multicolumn{2}{c}{\textbf{Average}} \\
\cmidrule(lr){3-4} \cmidrule(lr){5-6} \cmidrule(lr){7-8}
\cmidrule(lr){9-10} \cmidrule(lr){11-12} \cmidrule(lr){13-14} \cmidrule(lr){15-16}
& & p@1 $\uparrow$ & \#Token $\downarrow$ & p@1 $\uparrow$ & \#Token $\downarrow$ & p@1 $\uparrow$ & \#Token $\downarrow$ &
p@1 $\uparrow$ & \#Token $\downarrow$ & p@1 $\uparrow$ & \#Token $\downarrow$ & p@1 $\uparrow$ & \#Token $\downarrow$ &
p@1 $\uparrow$ & \#Token $\downarrow$ \\
\midrule

\multirow{6}{*}{\makecell[l]{Qwen3\\-8B}} & Vanilla & 71.67 & 17.6k & 69.58 & 15.8k & 38.75 & 19.4k & 42.38 & 19.5k & 4.59 & 13.3k & 38.50 & 19.4k & 44.24 & 17.5k \\
 & H2O & 60.00 & 17.8k & 70.00 & 18.3k & 46.67 & 21.5k & 35.00 & 20.5k & 4.39 & 15.3k & 36.25 & 20.7k & 42.05 \textcolor{red}{(-2.19)} & 19.0k \textcolor{red}{(+8.6\%)} \\
 & TC & 51.67 & 20.3k & 66.67 & 16.6k & 26.67 & 20.5k & 42.00 & 19.7k & 4.96 & 15.3k & 34.00 & 20.3k & 37.66 \textcolor{red}{(-6.58)} & 18.8k \textcolor{red}{(+7.4\%)} \\
 & No Train & 73.33 & 15.3k & 70.83 & 16.4k & 42.50 & 19.9k & 43.38 & 20.6k & \textbf{5.89} & 12.3k & 38.75 & 17.4k & 45.78 \textcolor{blue}{(+1.54)} & 17.0k \textcolor{blue}{(-2.9\%)} \\
 & Gemini & 71.67 & 13.7k & 68.75 & 15.2k & 44.17 & 18.4k & 43.63 & 17.6k & 5.10 & 11.8k & 40.00 & 17.8k & 45.55 \textcolor{blue}{(+1.31)} & 15.8k \textcolor{blue}{(-9.7\%)} \\
 & \ours{} & \textbf{75.00} & 13.0k & \textbf{72.50} & 13.7k & \textbf{49.58} & 16.4k & \textbf{45.88} & 15.4k & 5.38 & 9.9k & \textbf{40.50} & 14.3k & \textbf{48.14} \textcolor{blue}{(+3.90)} & 13.8k \textcolor{blue}{(-21.1\%)} \\
\midrule
\multirow{4}{*}{\makecell[l]{Qwen3\\-30B}} & Vanilla & 83.33 & 14.9k & 82.92 & 15.1k & 60.83 & 21.0k & 56.50 & 22.5k & 8.99 & 13.9k & 52.25 & 21.1k & 57.47 & 18.1k \\
 & No Train & 85.21 & 15.9k & \textbf{86.25} & 17.1k & 63.33 & 20.4k & 59.25 & 20.8k & 9.78 & 12.7k & 53.25 & 19.4k & 59.51 \textcolor{blue}{(+2.04)} & 17.7k \textcolor{blue}{(-2.2\%)} \\
 & Gemini & \textbf{87.92} & 15.4k & 84.17 & 15.9k & 65.42 & 19.0k & 59.75 & 19.7k & 9.59 & 12.8k & 54.00 & 20.4k & 60.14 \textcolor{blue}{(+2.67)} & 17.2k \textcolor{blue}{(-5.0\%)} \\
 & \ours{} & \textbf{87.92} & 14.5k & 85.42 & 14.0k & \textbf{70.42} & 19.1k & \textbf{62.25} & 18.2k & \textbf{9.82} & 11.3k & \textbf{58.00} & 18.1k & \textbf{62.30} \textcolor{blue}{(+4.83)} & 15.9k \textcolor{blue}{(-12.2\%)} \\
\midrule
\multirow{4}{*}{\makecell[l]{Qwen3\\-235B}} & Vanilla & 92.29 & 21.7k & 91.67 & 19.9k & \textbf{85.00} & 29.1k & 69.00 & 31.0k & 16.13 & 22.5k & 61.00 & 32.3k & 69.18 & 26.1k \\
 & No Train & 90.63 & 16.2k & 92.08 & 17.1k & 80.00 & 24.4k & 68.63 & 22.9k & 15.20 & 18.2k & 58.25 & 24.4k & 67.46 \textcolor{red}{(-1.72)} & 20.5k \textcolor{blue}{(-21.5\%)} \\
 & Gemini & \textbf{93.54} & 18.5k & 93.75 & 17.7k & 84.17 & 25.6k & \textbf{70.38} & 26.5k & 16.59 & 18.7k & 62.25 & 27.6k & 70.11 \textcolor{blue}{(+0.93)} & 22.4k \textcolor{blue}{(-14.2\%)} \\
 & \ours{} & 93.13 & 16.4k & \textbf{94.58} & 16.4k & 84.58 & 21.7k & 69.75 & 21.3k & \textbf{18.03} & 15.8k & \textbf{62.75} & 24.2k & \textbf{70.47} \textcolor{blue}{(+1.29)} & 19.3k \textcolor{blue}{(-26.1\%)} \\
\bottomrule
\end{tabular}
}
\end{table*}

\paragraph{Higher Accuracy with Deep Pruning:}
The most striking finding is the substantial boost in reasoning capability. On the Qwen3-8B benchmark, \ours{} achieves an average Pass@1 of \textbf{48.14\%}, outperforming vanilla inference (44.24\%) by \textbf{+3.9\%}. 
Crucially, this scaling is achieved by retaining \textit{less} rather than generating \textit{more}. We observe a distinct ``Deep Pruning'' phenomenon: \ours{} yields a significantly shorter average response length (13.8k) than the Gemini-2.5-Pro (15.8k), while surpassing its accuracy by \textbf{+2.3\%}.

Our qualitative investigation reveals the decisive factor: while Gemini often mistakenly removes useful clues—forcing the backbone to regenerate them—\ours{} precisely targets \textbf{true redundancy}. As illustrated in the case study (Figure~\ref{fig:case_study}), \ours{} maintains a noise-free reasoning chain without any observed regeneration. This proves the effectiveness of our reward mechanism: we successfully trained a module that knows exactly what to prune, reaching a precision level that even top-tier models struggle to achieve.


\paragraph{The Failure of Heuristic Compression:}
The results for H2O and TC are puzzling. Not only did they fail to improve accuracy, but they also failed to reduce the response length. A close inspection reveals the cause: \textbf{catastrophic degeneration}. By using heuristics for pruning, these methods disrupt the reasoning chain, causing the model to get stuck in repetitive loops. Ultimately, instead of making the model more intelligent, these heuristic methods render the reasoning process prone to crashing and produce even longer, redundant outputs.

\paragraph{Scalability:}
On the massive Qwen3-235B-A22B, \ours{} demonstrates its true value on challenging tasks like HLE and IMOAnswer. It boosts performance significantly (\textbf{11.7\% relative improvement} on HLE) while reducing context length by a massive \textbf{27.5\%}. This sends a clear message: \textbf{bigger is not always better if the context is cluttered}. 
For these giants to perform at their peak, the ability to \texttt{free()} redundant information is as vital as the capacity to store it.

\begin{wraptable}{r}{0.5\textwidth}
\centering
\vspace{-14pt}
\caption{Performance of \ours{} (Qwen3-8B Backbone).}
\label{tab:generalization}
\setlength{\tabcolsep}{2pt} 
\begin{tabular}{lcccc}
\toprule
\textbf{Method} & \textbf{BBH} & \textbf{MMLU-P} & \textbf{MMLU-S} & \textbf{GPQA} \\
\midrule
Qwen3-8B & 82.86 & 75.57 & 92.29 & 59.41 \\
\textbf{\ours{}} & 82.89 & 75.58 & 92.55 & \textbf{61.36} \\
\bottomrule
\end{tabular}
\end{wraptable}

\paragraph{Short-Reasoning Tasks:}
A common concern is whether pruning hurts general capabilities. Table~\ref{tab:generalization} allays this fear, showing \ours{} maintains parity across broad benchmarks. Since the backbone is frozen, the Free-Module is triggered significantly less often on short-reasoning tasks. As a result, \ours{} closely tracks the performance of the vanilla model without any degradation in general knowledge.

\begin{figure}[t]
  \centering
  \includegraphics[width=0.6\linewidth]{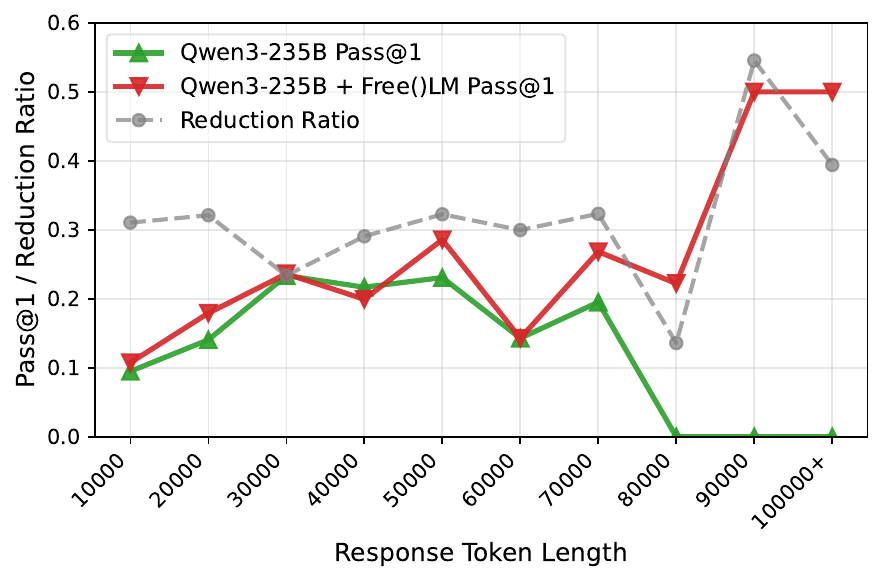}
  \caption{\textbf{Performance vs. Reasoning Length on HLE.} While standard Qwen3-235B-A22B suffers from Degradation on trajectories longer than 80k tokens, \ours{} exhibits a striking rebound in accuracy. On these cases, the Free-Module reduces the context length by $\sim$45\%, effectively mitigating context pollution.}
  \label{fig:length_analysis}
\end{figure}

\subsection{Cross-Model Generalization}

Following the superior performance of \ours{}, we investigate a critical question: is the capability of a trained Free-Module model-specific, or does it generalize across different architectures? To explore this, we first apply the Free-Module trained on Qwen3-8B to perform the \texttt{free()} operation for the much larger Qwen3-235B-A22B on IMOAnswer. Results in Table \ref{tab:cross-model} demonstrate that the 8B Free-Module achieves performance gains ($+1.5\%$) comparable to those of the 235B Free-Module ($+1.75\%$ as shown in Table \ref{tab:qwen3-performance}).

\begin{wraptable}{r}{0.5\textwidth} 
    \centering
    \vspace{-10pt} 
    \caption{\textbf{Cross-Model Generalization.} The 8B Free-Module enhances 235B/685B performance on IMOAnswerBench.}
    \label{tab:cross-model}
    
    \begin{tabular*}{\linewidth}{l @{\extracolsep{\fill}} c c}
        \toprule
        \textbf{Inference Setup} & \textbf{Pass@1} & \textbf{Token Reduction} \\
        \midrule
        Qwen3-235B & 61.00 & -- \\
        \textbf{+ Free-Mod (8B)} & \textbf{62.50} & \textbf{21.34\%} \\
        \midrule
        DeepSeek-V3.2 & 83.54 & -- \\
        \textbf{+ Free-Mod (8B)} & \textbf{85.87} & \textbf{45.99\%} \\
        \bottomrule
    \end{tabular*}
    \vspace{-10pt} 
\end{wraptable}

However, \textbf{a critical question remained}: is this generalization capability merely a byproduct of the shared model family? To challenge this hypothesis, we introduced a completely different architecture, \textbf{DeepSeek-V3.2-Speciale} \citep{deepseekv32}, into our evaluation. The results addressed this concern: even on this ``alien'' architecture, the Qwen-backboned 8B Free-Module improved Pass@1 by \textbf{2.3\%} while simultaneously slashing response tokens by \textbf{45.99\%}. This substantial improvement across distinct architectures suggests that the 8B-module has acquired a \textit{universal} context-cleaning capability. This cross-model generalization inspires a plug-and-play deployment strategy: serving the Free-Module as a \textbf{``Universal Context Pruning Service.''} In this paradigm, any backbone model can simply invoke this service upon reaching a trigger condition, pruning its context, and resume inference without requiring model-specific adaptation.




\subsection{Analysis on Reasoning Length}
\label{sec:exp_length}

Figure~\ref{fig:length_analysis} breaks down the performance on HLE. To isolate the impact of reasoning depth, we categorize test instances based on the token count of the response generated by the vanilla Qwen3-235B-A22B model. The results reveal a stark contrast:

\textbf{Reasoning models suffer from Degradation.} 
For manageable lengths, the vanilla model performs well. However, as trajectories extend beyond 80k tokens, the accumulated noise triggers a catastrophic degradation, and resulting in a complete loss of accuracy to 0\%.

\textbf{\ours{} recovers the ``Lost'' Intelligence.} 
In these same deep waters ($>$80k tokens), \ours{} exhibits a striking rebound, with accuracy surging back to $\sim$50\%. Although the limited sample size in this tail warns against claiming a definitive ``intelligence boost,'' the ability to \textit{sustain} reasoning capability is undeniable. 

The grey dashed line explains this recovery. On these ultra-long paths, the Free-Module aggressively compresses the context by 40\%--50\%. This operation successfully brings a massive 100k+ trajectory back down to the 40k--70k range, which is a ``sweet spot'' where the Qwen3-235B-A22B backbone operates most comfortably. We may simply conclude the result as: \textbf{with greater context comes a greater necessity for \ours{}.}


\subsection{System Performance}
Finally, we evaluate the engineering impact on Qwen3-235B-A22B using the HLE benchmark. We deploy the model across two 8$\times$H20 nodes using Tensor Parallelism (TP=16). The serving backend is \texttt{vLLM} (v0.8.5) on CUDA 12.6, configured with FlashAttention-2 and BF16 precision.

\begin{wraptable}{r}{0.48\textwidth} 
    \centering
    \small
    \vspace{-12pt} 
    \caption{\textbf{Efficiency Comparison.} Per-sample latency and KV cache memory usage for Qwen3-235B on HLE.}
    \label{tab:efficiency}
    
    \begin{tabular*}{\linewidth}{l @{\extracolsep{\fill}} c c c}
        \toprule
        \textbf{Metric} & \textbf{Base} & \textbf{\ours{}} & \textbf{$\Delta$} \\
        \midrule
        Latency (s) & 353.2 & 552.3 & +56.4\% \\
        KV (GB) & 6.14 & \textbf{3.34} & \textbf{-45.6\%} \\
        \bottomrule
    \end{tabular*}
    \vspace{-10pt} 
\end{wraptable}

We start with the cost: \ours{} incurs a 56\% increase in latency per sample. This overhead primarily stems from three sources: (1) the time spent decoding \texttt{free()} commands, (2) the re-prefilling latency incurred after executing a deletion, and (3) the regeneration of information if the model occasionally over-prunes.

However, the upside is substantial: we achieve a \textbf{45\% reduction} in KV cache usage (6.14 GB $\to$ 3.34 GB). In real-world serving, where memory bandwidth is often the bottleneck, this saving is critical. Furthermore, the current latency is not a hard limit. By adopting \textbf{Strategy 2} (KV Cache Pruning) to eliminate the re-prefilling step mentioned in (2), we estimate the overhead could be cut down to $\sim$20\%, although we have not yet implemented this strategy in vLLM.

\begin{figure*}[t]
  \centering
  \setlength{\fboxsep}{4pt}
  \setlength{\fboxrule}{0.4pt}

  \def\boxheight{130pt} 

  \begin{minipage}[t]{0.485\textwidth}
    \small\textbf{\ours{}: Stable Deletion}\vspace{2pt}

    \fbox{%
    \begin{minipage}[t][\boxheight]{0.975\textwidth} 
      \raggedright\scriptsize\ttfamily
      \linespread{0.9}\selectfont
      \textbf{[Current step]}\par
      \textcolor{black}{According to previous analysis, two non-empty intersections. However, the problem states three regions. Therefore, perhaps I made an error in considering Case 2 $\cap$ Case B?}\par
      \cotblock{red!12}{\cotdel{Wait, let me check again.\newline ...\newline Therefore, Case 2 $\cap$ Case B is empty.}}\par
      \textcolor{black}{Therefore, the only non-empty intersections are Case 1 $\cap$ Case B and Case 2 $\cap$ Case A. But the problem states three regions. So, where is the third region?}\par\vspace{4pt}

      \textbf{[Next step](continues after deletion)}\par
      \cotgen{Suppose we consider the entire plane $x + y + z = 75$. The inequalities divide it into regions. Since the problem says three regions, maybe there's another region where both inequalities hold? ...}\par
    \end{minipage}}
  \end{minipage}
  \hfill
  \begin{minipage}[t]{0.485\textwidth}
    \small\textbf{Gemini Deletion: Regeneration}\vspace{2pt}

    \fbox{%
    \begin{minipage}[t][\boxheight]{0.975\textwidth} 
      \raggedright\scriptsize\ttfamily
      \linespread{0.9}\selectfont
      \textbf{[Current step]}\par
      \textcolor{black}{Alternatively, since $\Gamma$ is the nine-point circle, maybe there are some symmetries?}\par
      \cotblock{red!12}{\cotdel{Maybe use coordinates\newline ...\newline find circumcircle of $D,E,F$}} 
      \par\textcolor{black}{Let me find the equation of the circle passing through these three points.}\par\vspace{4pt}

      \textbf{[Next step](re-generated)}\par
      \cotgen{Also, angle at B is $60^\circ$, which can be used to find coordinates.}\par
      \cotblock{blue!12}{\textcolor{blue}{Maybe use coordinates}\newline
      ...\newline
      \textcolor{blue}{find circumcircle of $D,E,F$}}\par
      \cotgen{Let me find the equation of the circle ...}\par
    \end{minipage}}
  \end{minipage}
\caption{Case study comparing \ours{} versus Gemini deletion. Deleted spans are shown in \textcolor{red}{red}, new generated content in \textcolor{ForestGreen}{green}, and re-generated content matching previous deletions in \textcolor{blue}{blue}. \ours{} (left) successfully prunes redundant reasoning; in contrast, Gemini (right) erroneously deletes critical logical anchors, forcing the subsequent reasoning model to re-generate the previously pruned context to restore the integrity of the reasoning chain.}
  \label{fig:case_study}
\end{figure*}

\subsection{Case Study} We qualitatively compare \ours{} and an ICL-based Gemini baseline using an AIME2425 example (Figure~\ref{fig:case_study}). \ours{} (left) demonstrates precise deletion: when the model begins redundant self-correction (denoted by \textcolor{red}{red}), the Free-Module removes it, allowing inference to resume (denoted by \textcolor{ForestGreen}{green}) without re-generating the pruned content. Conversely, Gemini (right) misjudges essential coordinate-setup as redundant. Its immediate regeneration (denoted by \textcolor{blue}{blue}) confirms the deletion was erroneous, as the backbone still required that information. This ``prune-then-regenerate'' cycle highlights Gemini’s inability to distinguish truly disposable context, resulting in wasted computation. Thus, it explains why \ours{} outperforms the Gemini-based Free-Module in Table \ref{tab:qwen3-performance}.












\section{Related Work}

\ours{} tackles context accumulation—a challenge where unbounded intermediate thoughts crowd out information needed for subsequent reasoning. We situate our approach at the intersection of KV cache compression, context window expansion, and CoT overthinking.

\subsection{KV Cache Compression}
The quadratic growth of the Key-Value (KV) cache is a primary bottleneck in LLM inference. To bound memory, heuristic eviction methods like StreamingLLM~\citep{streamingllm} and H2O~\citep{h2o} preserve "anchor" or high-attention tokens. SnapKV~\citep{snapkv} and PyramidKV~\citep{pyramidkv} refine this via saliency clustering and layer-wise adaptive budgets. Other strategies include reconstruction-based methods like KVzip~\citep{kvzip} and low-bit quantization such as KIVI~\citep{liu2024kivi}. However, lossy compression can induce "distraction"~\citep{kvpitfalls}, shifting attention to irrelevant content and degrading performance on precise reasoning tasks. Unlike these throughput-centric approaches, \ours{} implements logic-aware pruning. By training a Free-Module to detect semantic redundancy within reasoning traces, \ours{} improves efficiency and maintains reasoning quality.

\subsection{Long Context Window}
Expanding the effective context window is a major research frontier. Early architectural innovations modified positional encodings to extend context limits. ALiBi~\citep{alibi} replaces absolute positions with relative biases to improve length extrapolation. YaRN~\citep{yarn} and LongRoPE~\citep{longrope} extend RoPE via rescaling/interpolation, enabling million-token contexts and pushing theoretical limits beyond 2M tokens. System-level approaches such as Ring Attention~\citep{ringattention} distribute attention computation across GPUs using blockwise schemes, further increasing feasible context length. Training-based methods such as LongLoRA~\citep{longlora} adapt pre-trained models to longer contexts efficiently. Despite these advances, \citet{contextlengthhurts} show performance can degrade as context grows due to distraction, even with perfect retrieval; \citet{nolima} similarly reports accuracy decay beyond 32k tokens across diverse reasoning tasks. This is orthogonal but complementary to \ours{}: rather than expanding the window, we address the ``memory leak'' by teaching models to actively \texttt{free()} obsolete information.

\subsection{Overthinking}
While Chain-of-Thought (CoT)~\citep{wei2022cot} improves model reasoning, it can induce overthinking: excessive, redundant steps that consume context without improving the final answer~\citep{overreasoning}. This is especially acute in multi-step math, where models may explore dead ends, repeat computations, or over-verify. The rise of o1-style long-reasoning models intensifies this issue, with 10k--100k+ tokens for a single problem; \citet{donotthinkmuch} shows redundancy can appear even in trivial arithmetic. Prior work mitigates overthinking via budgeted generation \citep{tale}, pruning based on token importance \citep{thinkclearly}, or stopping rules such as entropy monitoring~\citep{cumulativeentropy}. Other methods optimize conciseness through training, e.g., O1-Pruner~\citep{o1pruner} uses reinforcement learning to reward shorter correct reasoning, and DeepCompress~\citep{deepcompress} employs adaptive length rewards based on problem difficulty. Unlike methods that primarily control output length, \ours{} enables dynamic context cleaning during inference through a trainable Free-Module that \texttt{free()}s redundant reasoning steps, keeping the workspace efficient throughout generation.

\section{Conclusion}
We demonstrate that the prevailing ``malloc-only'' paradigm—where context is treated as a passive, append-only buffer—is fundamentally unsustainable for long-horizon reasoning. In standard LLMs, the continuous accumulation of redundant tokens eventually overwhelms the model, leading to the total collapse of reasoning performance. To break this cycle, we introduce \textbf{\ours{}}, completing the memory management cycle with the missing \texttt{free()} operation.

Our work makes three primary contributions. First, we provide the first systematic evidence that active context pruning directly enhances reasoning performance across all model scales, from \textbf{8B to 685B}. Second, we show that our plug-and-play \textbf{Free-Module} effectively maintains a noise-free state, yielding a \textbf{3.3\%} average gain over leading baselines and setting a new \textbf{SOTA} on IMOanswerBench. Third, and most crucially, we demonstrate that \ours{} solves the failure of long-reasoning: while standard models like Qwen3-235B-A22B drop to \textbf{0\% accuracy} on complex tasks, \ours{} restores performance to \textbf{28\%} by precisely pruning redundancy.

Ultimately, \ours{} suggests a shift in the scaling laws of test-time compute. The path to long-horizon intelligence lies not merely in expanding the context window, but in mastering the \textbf{art of forgetting}. This transition from ``malloc-only'' to ``malloc + free'' provides the foundation for the next generation of efficient, self-sustaining reasoning agents.

\bibliographystyle{unsrt}
\bibliography{custom}

\appendix

\clearpage
\section{Experimental Details}

\subsection{Prompts}
\label{sec:prompts}

\begin{figure*}[ht]
\centering

\noindent\fcolorbox{black!20}{cyan!10}{%
\begin{minipage}{0.98\textwidth}
\small
\textbf{System Prompt:}

\vspace{0.1em}
\textit{Role: AI Reasoning Analyst}

\vspace{0.1em}
Your task is to act as an AI Reasoning Analyst. You will be given a Chain-of-Thought (CoT) reasoning and your goal is to identify and mark for deletion any paragraphs from the CoT reasoning that are redundant or irrelevant.

\vspace{0.1em}
\textbf{Instructions:}
\begin{enumerate}
\vspace{0.2cm}
\item \textbf{Analyze the CoT Reasoning:} Carefully read and understand the reasoning, context, and goals of the CoT reasoning.
\item \textbf{Identify Redundant or Irrelevant Paragraphs:} A paragraph is considered redundant or irrelevant if it does not directly support or inform the overall reasoning process. This could include tangential thoughts, corrected errors, or superseded lines of reasoning.
\item \textbf{Mark for Deletion:} For every irrelevant paragraph you detect, generate a JSON object that uniquely identifies it. Each object must include a \texttt{prefix} and a \texttt{suffix} extracted from the paragraph.
\\item \textbf{Format the Output:} Your final output must be a single JSON object containing a list of these \texttt{prefix}/\texttt{suffix} objects.
\item \texttt{<DELETED>} in the CoT reasoning indicates that the content has already been removed.
\item It is possible that you don't need to delete any paragraphs in the CoT reasoning.
\end{enumerate}

\vspace{0.1em}
\textbf{Constraints:}
\begin{itemize}
\item If no paragraphs are redundant or irrelevant, output an empty list: \texttt{[]}.
\item If multiple paragraphs are redundant, include a separate JSON object for each one.
\item Do not delete the first or the last paragraph in the CoT reasoning.
\end{itemize}
\end{minipage}%
}

\vspace{0.5em} 

\noindent\fcolorbox{black!20}{green!10}{%
\begin{minipage}{0.98\textwidth}
\small
\textbf{User Prompt:}

\vspace{0.1em}
\texttt{\#\#\# CoT Reasoning:}\\
\texttt{<PREVIOUS\_COT>}
\end{minipage}%
}

\caption{\textbf{Deletion Prompts for Free()LM.} The system prompt (top) provides the operational logic for identifying redundant reasoning steps, while the user prompt (bottom) delivers the actual CoT context for the agent to process.}
\label{fig:prompts}
\end{figure*}

\begin{figure*}[t]
\centering

\noindent\fcolorbox{black!20}{orange!10}{%
\begin{minipage}{0.98\textwidth}
\small
\textbf{Response Schema (JSON):}

\vspace{0.2em}
\texttt{\{}\\
\texttt{\hspace*{1.5em}"type": "array",}\\
\texttt{\hspace*{1.5em}"items": \{}\\
\texttt{\hspace*{3.0em}"type": "object",}\\
\texttt{\hspace*{3.0em}"properties": \{}\\
\texttt{\hspace*{4.5em}"prefix": \{}\\
\texttt{\hspace*{6.0em}"type": "string",}\\
\texttt{\hspace*{6.0em}"description": "Beginning text of paragraph to delete."}\\
\texttt{\hspace*{4.5em}\},}\\
\texttt{\hspace*{4.5em}"suffix": \{}\\
\texttt{\hspace*{6.0em}"type": "string",}\\
\texttt{\hspace*{6.0em}"description": "Ending text of paragraph to delete."}\\
\texttt{\hspace*{4.5em}\}}\\
\texttt{\hspace*{3.0em}\},}\\
\texttt{\hspace*{3.0em}"required": ["prefix", "suffix"]}\\
\texttt{\hspace*{1.5em}\},}\\
\texttt{\hspace*{1.5em}"description": "A list of prefix/suffix pairs that uniquely identify paragraphs that are redundant or irrelevant and should be deleted."}\\
\texttt{\}}
\end{minipage}%
}

\caption{\textbf{Structured Output Schema for Free()LM.} The JSON schema enforces a consistent prefix/suffix format, ensuring that redundant or irrelevant reasoning steps identified by the agent can be programmatically parsed and removed from the context.}
\label{fig:response_schema}
\end{figure*}

The prompts utilized in this study are detailed in Figure \ref{fig:prompts}. We employ a consistent prompt for both training corpus preparation and alignment evaluation. To ensure a structured response format, we implement a JSON schema to guide the Free-Module's output, as illustrated in Figure \ref{fig:response_schema}. Both vLLM \citep{vllm} and Gemini natively support this structured output functionality.

\subsection{Training Details}

Since \ours{} is trained to perform deletion over intermediate reasoning produced by backbone models that do not expose explicit reasoning traces as supervision, we disable the \emph{thinking mode} in the chat template during prompt construction. This design ensures that the Free-Module operates solely on the visible model-generated reasoning, matching the inference-time setting.

We train the Free-Module on the Qwen3-8B backbone using TRL (v0.19.1) \citep{trl}, while the larger Qwen3-30B-A3B and Qwen3-235B-A22B variants are trained with Megatron-LM (v0.14.1) \citep{megatron}. All models are trained for 5 epochs with a learning rate of $1 \times 10^{-5}$ and a global batch size of 16. To optimize memory efficiency and throughput, we use the AdamW optimizer together with FlashAttention-2 for attention computation. We employ DeepSpeed ZeRO Stage~3 for Qwen3-8B and Stage~2 for Qwen3-30B-A3B and Qwen3-235B-A22B, balancing memory savings and communication overhead across model scales. All training runs are conducted using BF16 precision.

For parameter-efficient fine-tuning, we apply LoRA to all linear layers with rank $r=128$, scaling factor $\alpha=256$, and dropout rate 0.1. The backbone model parameters remain frozen throughout training; only the Free-Module parameters are updated.

\subsection{Evaluation Details}

For prefilling, we append the model’s previously generated response to the end of the prompt after applying the chat template, and re-run the prompt through the model to resume generation from the refined context. This procedure ensures that subsequent tokens are generated conditionally on the updated reasoning trace, rather than restarting inference from scratch.

In practice, we leverage vLLM’s support for efficient prefilling and KV cache reuse. When the context is modified by a \texttt{free()} operation, we reuse the cached key–value states corresponding to the unchanged prefix, and only re-prefill the altered suffix. This avoids recomputing attention for the entire context and significantly reduces the overhead of resuming inference.

\end{document}